%% file: main.tex
\documentclass{article}

\usepackage{arxiv}
\usepackage{subfiles}

\usepackage[utf8]{inputenc} 
\usepackage[T1]{fontenc}    
\usepackage{hyperref}       
\usepackage{url}            
\usepackage{booktabs}       
\usepackage{amsfonts}       
\usepackage{nicefrac}       
\usepackage{microtype}      
\usepackage{lipsum}
\usepackage[dvipsnames,svgnames]{xcolor}
\usepackage{graphicx} 

\usepackage{subcaption}

\usepackage{amsmath}
\usepackage{algpseudocode}
\usepackage{algorithm}

\algnewcommand\algorithmicforeach{\textbf{for each}}
\algdef{S}[FOR]{ForEach}[1]{\algorithmicforeach\ #1\ \algorithmicdo}

\title{Offline Grid-Based Coverage path planning for guards in games}

\author{
  Wael Al Enezi\\
  Department of Computer Science\\
  McGill University\\
  Montr\'eal, Canada \\
  \texttt{wael.alenezi@mail.mcgill.edu} \\
   \And
Clark Verbrugge \\
  Department of Computer Science\\
  McGill University\\
  Montr\'eal, Canada \\
  \texttt{clump@mcgill.ca} \\
}

\begin{document}
\maketitle

\subfile{sections/abstract}


\subfile{sections/introduction}

\subfile{sections/related_work}

\subfile{sections/methodology}

\subfile{sections/results}

\subfile{sections/conclusions}

\bibliographystyle{unsrt}  

\bibliography{main.bib}

\end{document}

%% file: sections/abstract.tex
\begin{abstract}
  Algorithmic approaches to exhaustive coverage have application in video games, enabling automatic game level exploration.  Current designs use simple heuristics that frequently result in poor performance or exhibit unnatural behaviour. In this paper, we introduce a novel algorithm for covering a 2D polygonal (with holes) area.  We assume prior knowledge of the map layout and use a grid-based world representation. Experimental analysis over several scenarios ranging from simple layouts to more complex maps used in actual games show good performance. This work serves as an initial step towards building a more efficient coverage path planning algorithm for non-player characters.

\end{abstract}

%% file: sections/introduction.tex
\section{Introduction}
Coverage path planning (CPP) is the process of planning a path for an agent that traverses across all areas of a specific location. One of the most challenging aspects of this problem is to optimize the planned path of the agent to cover the area of interest \cite{kuo2011pneumatic}. CPP can be categorized into two types: off-line or online \cite{choset2001coverage}. In off-line coverage, the environment layout is known to the agent, while in on-line coverage the agent has no prior knowledge of the environment. This problem can be approached by several methods. In this paper we use a grid-based approach to implement a novel algorithm to cover an area of interest. Our algorithm is easy to implement, but also guaranteed to solve any area of interest. The algorithm does not guarantee an optimal solution, but it provides the basis for extending the coverage to multiple agents without additional complexity.

In the next section we introduce the related work in the field of coverage path planning, then we describe the methodology of our work. After that, we describe the results the algorithm achieved on selected scenarios. Lastly, we draw our conclusions and future work.

%% file: sections/related_work.tex
\section{Related Work}
Coverage path planning has been well studied in the field of robotics for a wide variety of applications, such as vacuum cleaning robots \cite{yasutomi1988cleaning}, lawnmowers \cite{cao1988region}, unmanned aerial vehicle (UAV) surveillance \cite{basilico2015deploying}, etc.  The literature includes several methods for CPP, but the methods heavily depend on the representation of the space, either relying on continuous space representations, such as using convexity, or on discrete, grid-based models.

\paragraph{Convex Decomposition} decomposes the traversable space into simple, non-overlapping convex regions. After the decomposition, solving the problem of coverage path planning relies on finding the shortest path to visit these regions along with the coverage pattern for each region. Examples of appropriate coverage patterns are zigzag \cite{choset1998coverage}, spiral \cite{balampanis2017spiral}. Once the space has been decomposed into convex sub-regions, an adjacency graph can be used to represent the decomposed space. In the graph, the nodes are the convex regions and the edges between the nodes represent the adjacency between the convex regions. The agent uses this graph to plan the shortest path to traverse all convex regions and recursively the shortest coverage pattern for each region.

\paragraph{Grid-Based} approaches decompose the space into a grid of uniformly distributed nodes \cite{moravec1985high}. Each node can be traversable or non-traversable. In this representation, CPP is accomplished by traversing all reachable nodes, associating either a binary or probabilistic \cite{elfes1987sonar} property to coverage. Grid-based methods are resolution-dependent, and their completeness relies on the granularity of the grid. Grid-based representation is one of the simplest methods to use. However, for larger spaces, this method becomes computationally expensive compared to other methods \cite{thrun1998learning}.

%% file: sections/methodology.tex
\section{Methodology}

In this section, we describe the space representation method chosen for our algorithm. After that, we explain how our algorithm accomplishes coverage path planning to an area of interest.

\subsection{Space Representation}
\begin{figure}[h]
    \centering
    \includegraphics[ width = 0.7\textwidth]{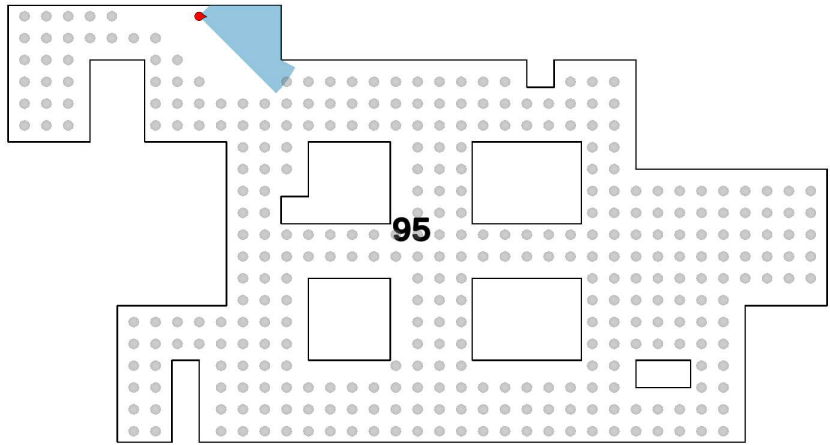}
    \caption{An example of the simulation. The map is taken from the game ``Metal Gear Solid''. The darkest nodes are unseen traversable nodes. Once an unseen node is seen it becomes transparent. The lighter nodes are untraversable. The number on the center indicates the percentage of unseen nodes.}
    \label{fig:grid_example}
\end{figure}

For our method, we use a 2D grid-based representation; The space is discretized into a grid of nodes, and each node has three possible states: untraversable, seen and unseen. The grid connectivity is based on von Neumann neighborhood where the agent moves by taking a step to the adjacent nodes in four directions; up, down, left and right. The agent's field of view is represented by a cone with a limited range and angle; other FoV models are also possible. Once an unseen node falls within the vision cone that node becomes seen. The condition to consider a scenario to be successfully finished is when all the traversable nodes are seen. Figure \ref{fig:grid_example} shows an example of the simulation.

\subsection{RippleFront Algorithm}
For our approach, we based our implementation on the WaveFront coverage algorithm \cite{shivashankar2011real}. We refer to it as the ``RippleFront algorithm.'' It works by propagating the distance from the unseen nodes through the seen nodes. The propagation procedure is constantly executed every fixed time interval. 

Algorithm \ref{alg:rippleFront} shows the pseudo-code for the RippleFront where $N$ is the list of nodes in the grid, $D(n)$ is the Manhattan distance to the closest unseen node. At every iteration, $D(n)$ is updated by assigning the $Min(D(l))$ where $l$ is an adjacent to $n$. $D(n)$ value will be updated by adding the value of $D(n_i) + 1$, where $n_i$ is the neighbor node with the shortest distance from the closest unseen node. The added $1$ is the step cost from $n$ to $n_i$.

\begin{algorithm} \caption{RippleFront Algorithm} \label{alg:rippleFront}
\begin{algorithmic}[1]
\Require $N$, the list of nodes in the grid. 
\State Initialize $D(n) = 0$, for all $n \in \mathcal N$
\Repeat
\ForEach {$n \in \mathcal N $}
\If{$IsSeen(n)$}
\State $d_o \gets D(n)$
\State $D(n) \gets Get Min Distance To Unseen In Neighbourhood(n) $
\If {$d_o != D(n) +1$} 
\State  $D(n) \gets D(n) + 1$
\EndIf
\EndIf
\EndFor
\State \Until All Seen
\end{algorithmic}
\end{algorithm}

The RippleFront focuses on creating an overall road map that makes it trivial for the agent to decide the direction of its next step to discover unseen locations. However, the coverage path is not guaranteed to be optimal due to the greedy decision process the agent adopts, which is simply to face and step towards the node closest to an unseen node.  Nevertheless, this approach provides a guarantee to solve the map.

%% file: sections/results.tex
\section{Results}

\begin{figure}[h]
\centering
\begin{subfigure}[b]{.4\textwidth}
  \centering
  \includegraphics[width=.8\linewidth]{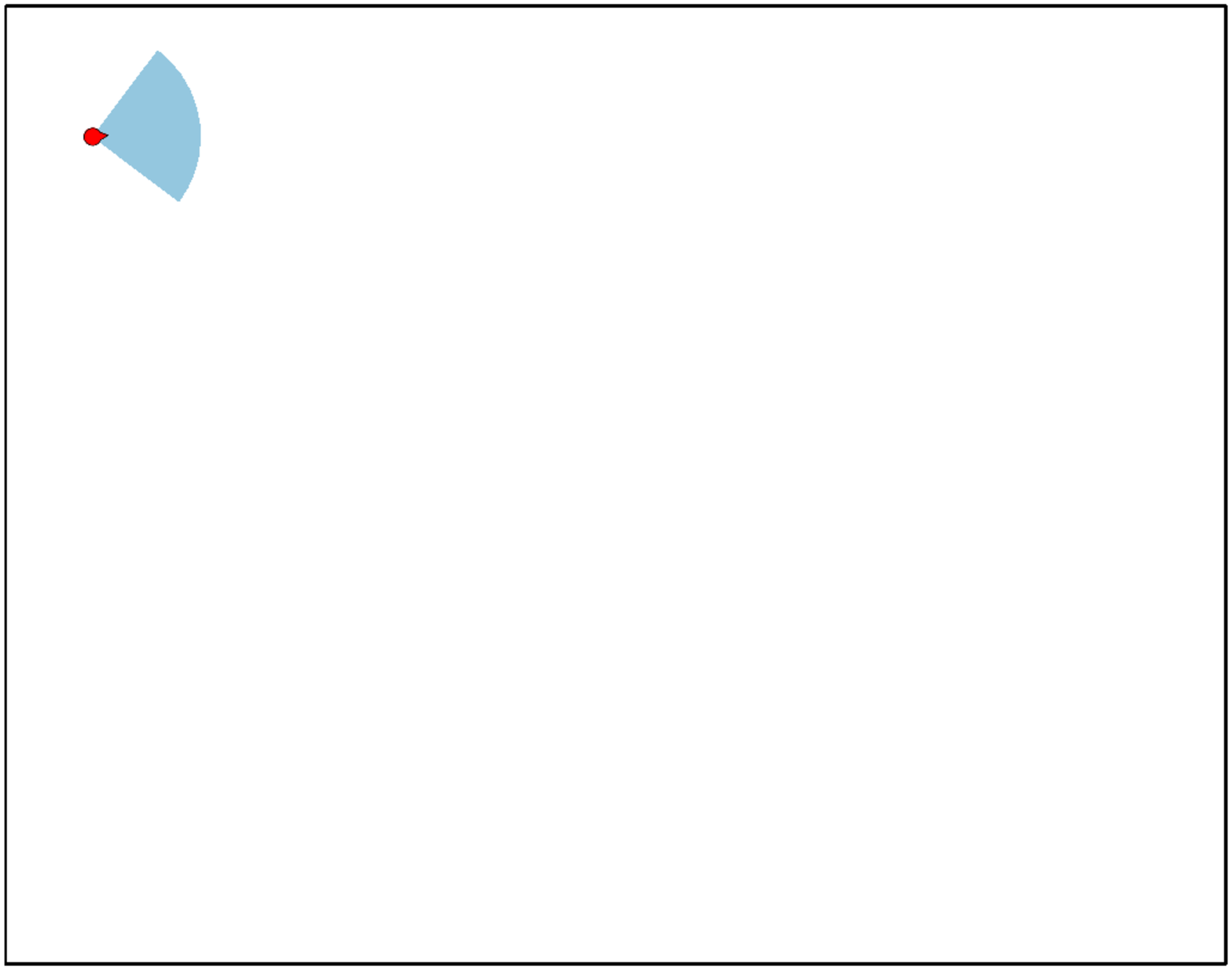}
  \caption{Square}
  \label{fig:square_screenshot}
\end{subfigure}
\begin{subfigure}[b]{.4\textwidth}
  \centering
  \includegraphics[width=.8\linewidth]{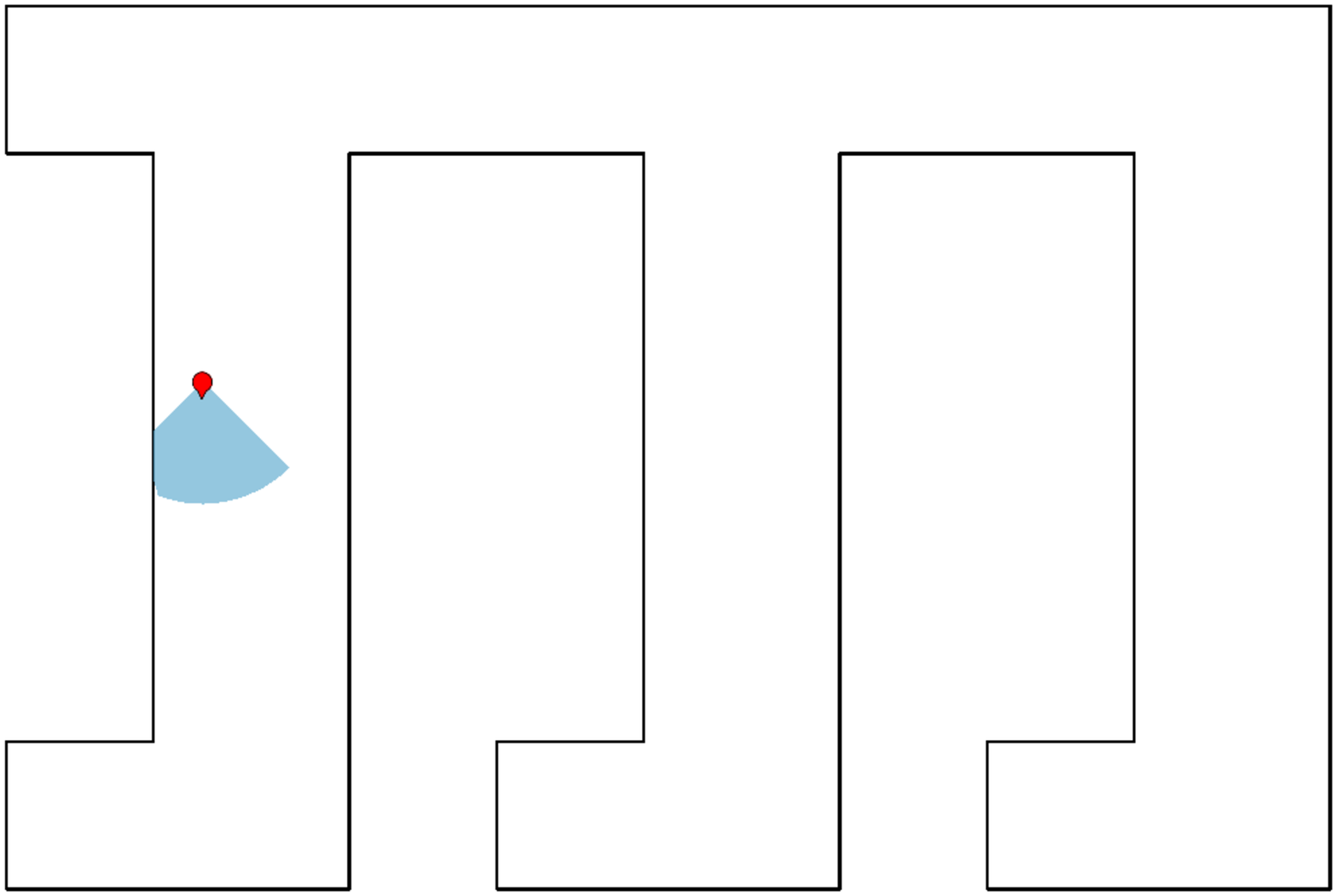}
  \caption{Passage ways}
  \label{fig:hooks_screenshot}
\end{subfigure}
\begin{subfigure}[b]{.4\textwidth}
  \centering
  \includegraphics[width=.8\linewidth]{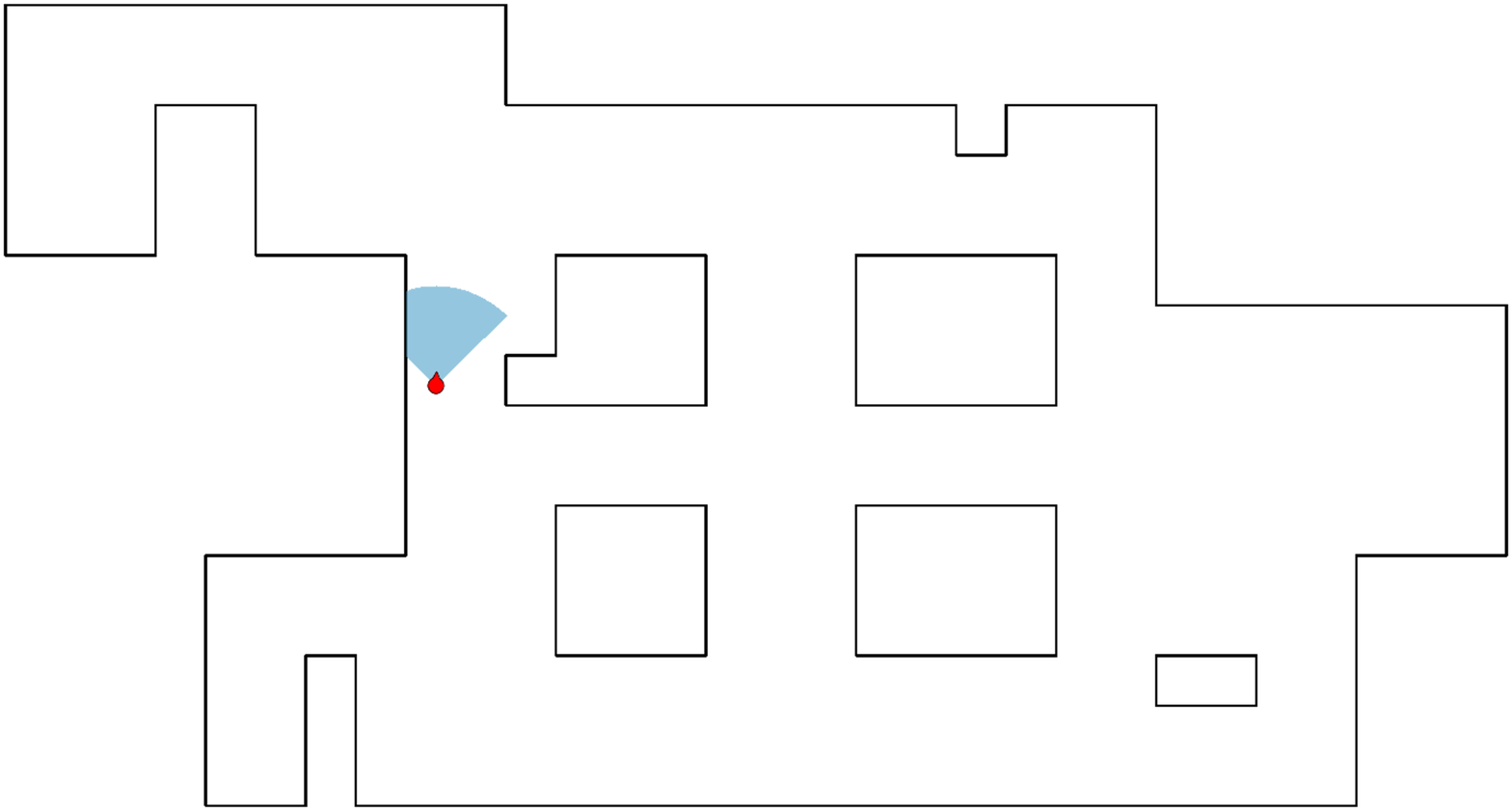}
  \caption{Metal Gear Solid Docks}
  \label{fig:mgsDocks_screenshot}
\end{subfigure}
\begin{subfigure}[b]{.4\textwidth}
  \centering
  \includegraphics[width=.8\linewidth]{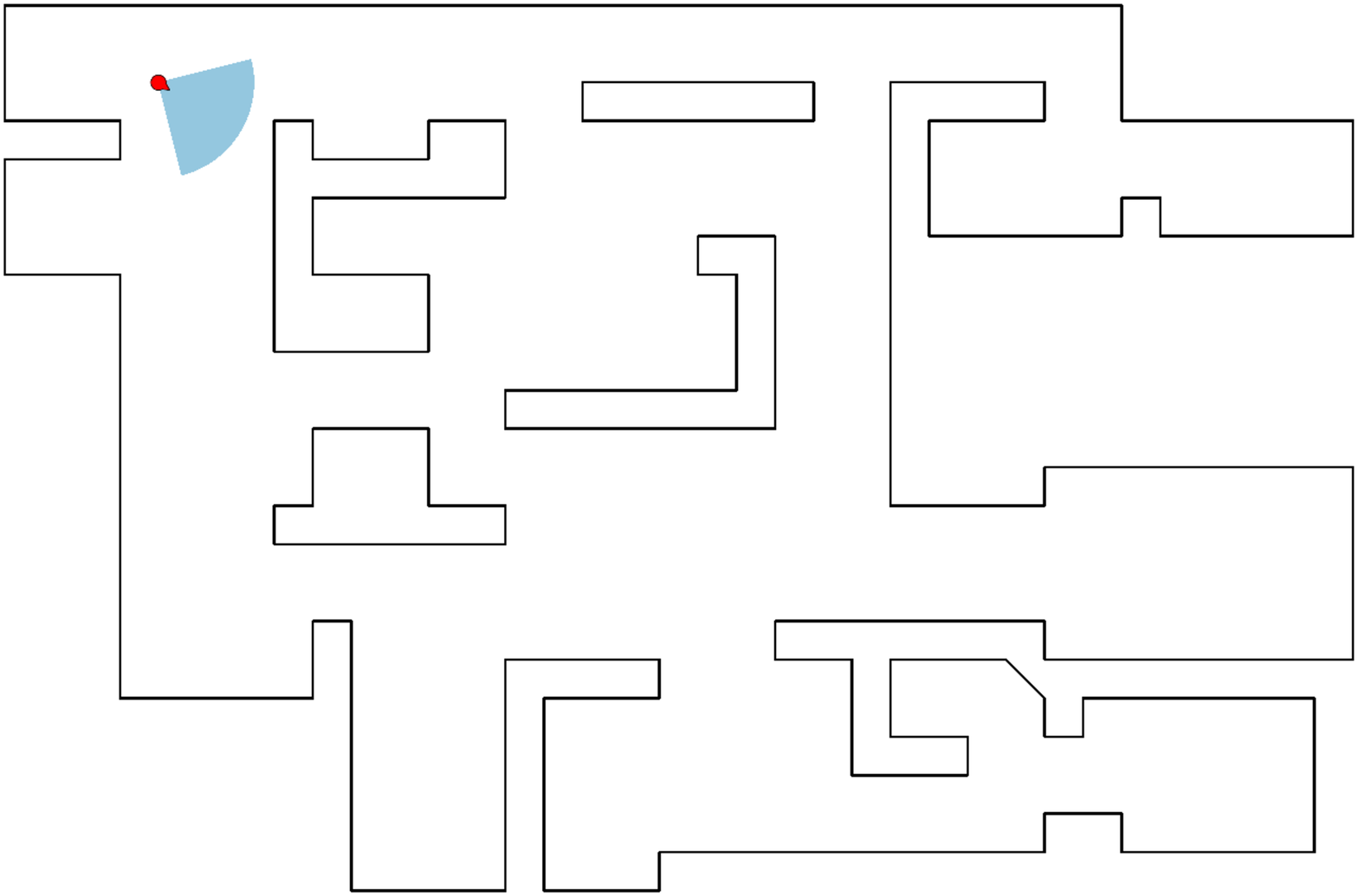}
  \caption{King Snurre's Hall}
  \label{fig:kingsrrul_screenshot}
\end{subfigure}
\caption{Maps used for the simulation.}
\label{fig:scenarios}
\end{figure}

We ran our simulation using Unity3D game engine \footnote{www.unity.com}. We tested the algorithm on 4 scenarios with different complexities. Figure \ref{fig:scenarios} shows the layout of the maps used for our simulations. The maps have different complexities but they all have the same grid size. 

\begin{figure}[h]
    \centering
    \includegraphics[width=0.4\textwidth]{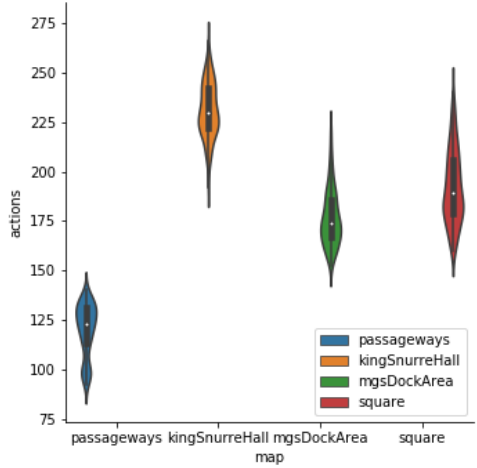}
    \caption{The performance for 200 episodes for each of the 4 scenarios.}
    \label{fig:over}
\end{figure}

To measure the agent's performance, we used the number of steps the agent required to cover a map. The length of a step is the distance between two adjacent nodes, vertically or horizontally. To assess the overall performance, We logged the total steps the agent took for each episode and ran 200 episodes for each scenario. For every episode the agent had a random starting position and tasked with covering the overall map.  Figure \ref{fig:over} shows the performance of our algorithm on the 4 scenarios. While all scenarios have the same grid size, each scenario has a different number of traversable nodes. This led to the difference in performance between these scenarios.

%% file: sections/conclusions.tex
\section{Conclusions \& Future work}

In this paper, we introduced a novel offline grid-based algorithm for coverage path planning. Our algorithm is simple to implement and guaranteed to solve any scenario. This is of course preliminary work.  Our future goals include conducting performance comparisons with other, existing algorithms or heuristics, considering computation time as well as number of steps.  We are also interested in extending the design to the context of multiple agents, where communicating agents cooperate to perform concurrent coverage.

\section{Acknowledgement}

This work was supported by the Natural Science and Engi-neering Research Council of Canada, Computing Hardware for Emerging Intelligent Sensing Applications (COHESA).